\documentclass[letterpaper, 10 pt, conference]{ieeeconf}  

\IEEEoverridecommandlockouts                              

\usepackage{graphicx} 
\usepackage{algorithm}
\usepackage{algorithmic}
\usepackage{amsmath} 
\usepackage{amsfonts}
\usepackage{cite}
\DeclareMathOperator*{\argmax}{\arg\!max}

\title{\LARGE \bf
Hindsight Reward Tweaking via Conditional Deep Reinforcement Learning
}

\author{Ning Wei$^{1}$, Jiahua Liang$^{1}$, Di Xie$^{1}$ and Shiliang Pu$^{1}$
\thanks{$^{1}$N. Wei, J. H. Liang, D. Xie, and S. L. Pu is with Hikvision Research Institute, 
        Hangzhou 310051, China
        {\tt\small \{weining5, liangjiahua, xiedi, pushiliang\}@hikvision.com}}%
}

\begin{document}

\maketitle
\thispagestyle{empty}
\pagestyle{empty}

\begin{abstract}

Designing optimal reward functions has been desired but extremely difficult in reinforcement learning (RL). When it comes to modern complex tasks, sophisticated reward functions are widely used to simplify policy learning yet even a tiny adjustment on them is expensive to evaluate due to the drastically increasing cost of training. To this end, we propose a hindsight reward tweaking approach by designing a novel paradigm for deep reinforcement learning to model the influences of reward functions within a near-optimal space. We simply extend the input observation with a condition vector linearly correlated with the effective environment reward parameters and train the model in a conventional manner except for randomizing reward configurations, obtaining a hyper-policy whose characteristics are sensitively regulated over the condition space. We demonstrate the feasibility of this approach and study one of its potential application in policy performance boosting with multiple MuJoCo tasks.

\end{abstract}

\section{INTRODUCTION}

As a classic approach to solve intelligent decision-making problem, reinforcement learning [1] has been on its way to revive with the development of deep learning technology in the last decade [2][3]. RL algorithms relies on reward functions to perform well. Despite the recent efforts in marginalizing hand-engineered reward functions [4][5][6] in academia, reward design is still an essential way to deal with credit assignments for most RL applications. [7][8] first proposed and studied the optimal reward problem (ORP). Later [9] reported that a somehow bounded agent can hardly achieve best performance under the direct guidance of the designer's goals yet well-designed alternative reward functions enable better and faster learning.

A general formulation of reward functions takes the form $ r\left(s,a,{s}'\right)=\omega^{T}\phi\left(s,a,{s}'\right) $, where $\omega$ is a scalar vector composed of encouraging/discouraging reward items and $\phi$ is a vector of predefined indicator features dependent on states $s$, ${s}'$, and the action $a$. With $\phi$ fixed, ORP seeks to find the optimal $\omega$ that leads to RL policies maximizing the given fitness functions. One major difficulty in reward design is the lack of instant feedback mechanism from $\omega$ to its actual effect due to the inherent inefficiency of RL algorithms. Most existing ORP-oriented approaches rely on sufficient training to provide good indicators for further improvement on $\omega$ [7][8][9][10][11][12][13][14][15]. However, the scalability of these approaches are undefined since they are only verified with small tasks.

\begin{figure}[t]
\centering
\includegraphics[width=1.0\columnwidth]{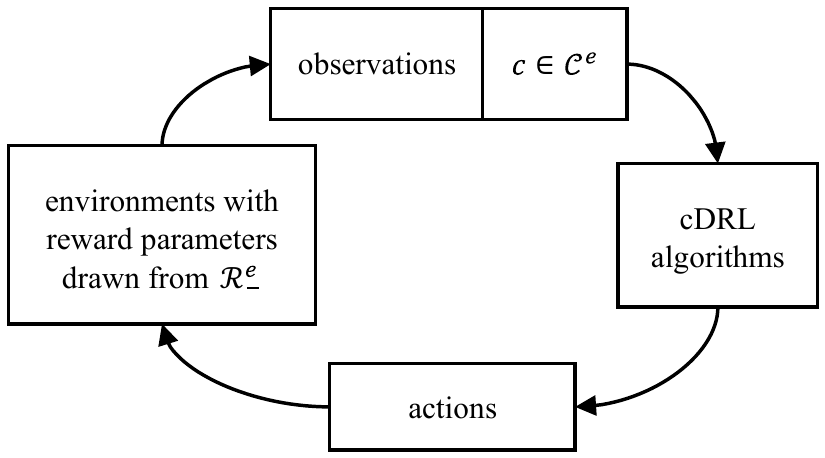} 
\caption{High-level overview of the proposed conditional deep reinforcement learning approach.}
\label{fig1}
\end{figure}

In this paper, we propose a new scalable approach named conditional deep reinforcement learning (cDRL). Instead of optimizing reward functions directly, we leverage the representation power of deep neural networks to model their influences on RL policies. As illustrated in Fig. 1, by extending the input observation with a condition linearly correlated with the effective reward parameters and training the model with corresponding featured examples, we expect deep RL algorithms to learn policies sensitive to this condition while being able to adapt behaviors according to the underlying long-period preferences. This approach is time \& resource efficient in the sense that it only makes tiny modifications to the frameworks and training processes of standard deep RL algorithms without requiring extra computing resource or prolonged training, so it can be easily applied to any large-scale tasks in a plug and play fashion.

Once a cDRL agent is trained, the input condition could solely act as a control panel to tweak the policy's characteristics in a totally hindsight perspective since the consequent effects can be readily measured without any further training. Despite the potential modeling inaccuracy on reward influences, which also should be well realized, cDRL indeed alleviates the dilemma in reward design as long as the policy stays sensitive to the input condition and their asymptotic high-level interaction mechanism is properly learned. Given the convenience introduced by cDRL, a straightforward application for it is to perform hindsight policy boosting with respect to fitness functions given by the designer. We validate this potential with multiple experiments in section 4.

\section{CONDITIONAL DEEP REINFORCEMENT LEARNING}

\subsection{Problem Set-Up}

When facing a new RL task, we need to figure out a group of indicator features based on observations, actions and history in accordance with the designer's goal. These features, either scalar or binary, should be highly expressive and correlated with the intrinsic logic behind the intended behaviors.

\textbf{Assumption 1.} \textit{For any specific task domain, we assume that all reward-related indicator features are predefined and well capable of conveying the designer's goal.}

Moreover, we also need to set a scalar vector, i.e. the reward parameters, as the weights of these indicator features, which is the core procedure for reward design in ORP. In RL, parallel vectors of reward parameters normally have equivalent effects. To tackle this redundancy, we select the first element as anchor\footnote{Actually any element in reward parameters can be set as anchor, we here choose the first element to make (1) concise.} with constant value $\xi$ and study the others' influences when they are varying with respect to it. This dimension reduced parameter space is denoted as $\mathcal{R}_{-}$ to distinguish from the whole space $\mathcal{R}$. Since we are only interested in the near-optimal region of $\mathcal{R}_{-}$, another assumption similar to [15] is made below.

\textbf{Assumption 2.} \textit{Each non-anchor reward parameter is assigned with a reasonable range according to the anchor based on available domain knowledge so that the sampled combinations are likely to lead to high true utility behaviors as desired.}

Essentially, these ranges delimit a subspace $\mathcal{R}_-^e$ of interest within $\mathcal{R}_-$. We argue that this is pragmatically much easier than setting exact optimal values for reward parameters. Hence our target now is to model the underlying interaction mechanisms among different dimensions of $\mathcal{R}_-^e$, which provides the opportunity to perform hindsight reward tweaking and achieve better performance than standard RL algorithms with hand-designed rewards.

\begin{figure}[t]
\centering
\includegraphics[width=1.0\columnwidth]{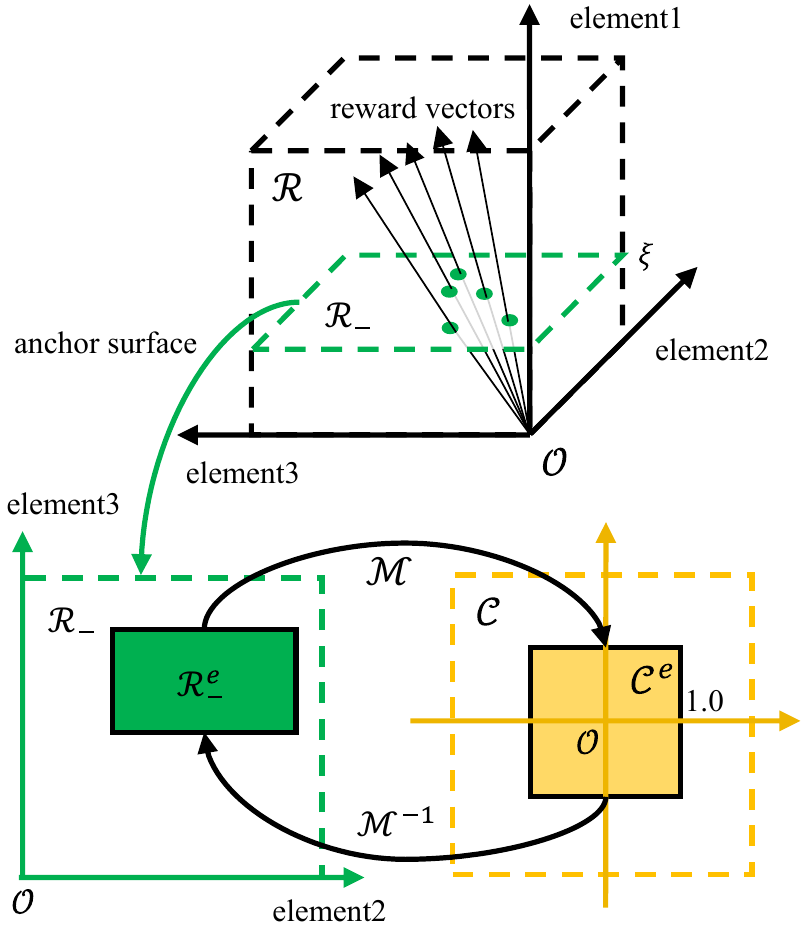} 
\caption{Illustration of reward-related space definitions and transformations in cDRL. For a RL task with N reward parameters, we settle the first element as anchor and the consequent hypersurface of dimension N-1 forms the non-redundant reward parameter space $\mathcal{R}_-$ to study. A near optimal subspace $\mathcal{R}_-^e$ is defined with domain knowledge and projected to $\mathcal{C}^e$ by a normalization operation with affine transformation $\mathcal{M}$. During cDRL training, reward parameters are sampled from $\mathcal{R}_-^e$ and output as their correspondences in $\mathcal{C}^e$ with observations.}
\label{fig2}
\end{figure}

To facilitate cDRL training for a certain domain, we need instance environments with randomized reward parameters uniformly drawn from $\mathcal{R}_-^e$ while outputting these parameters along with original observations at every step. Note that different dimensions of $\mathcal{R}_-^e$ may vary drastically in numerical scales, which increases learning difficulty. We thus introduce an adaptive affine transformation $\mathcal{M}:\mathcal{R}_-\rightarrow \mathcal{C}$ which simply maps each dimension of $\mathcal{R}_-^e$ to [-1,1] in $\mathcal{C}$. We name $\mathcal{C}$ as the condition space and denote the normalized subspace corresponding to $\mathcal{R}_-^e$ as $\mathcal{C}^e$(see Fig. 2). All reward parameters within $\mathcal{R}_-^e$ are projected to $\mathcal{C}^e$ before they are concatenated and output with the original observations. Then the stepwise reward function can be reformulated as:

\begin{equation}
    r_c\left(s,a,s'\right)=\left[\xi;\mathcal{M}^{-1}\left(c\right)\right]^T\phi\left(s,a,s'\right)
\end{equation}

\noindent where $c\in \mathcal{C}^{e}$, $\mathcal{M}^{-1}$ uniquely maps $c$ to a valid reward parameter vector within $\mathcal{R}_-^e$ which, combined with the anchor $\xi$, forms a whole group of weights for indicator features.

\subsection{CDRL Algorithms}

When applying deep RL to a certain domain, different reward parameters produce different returns and thus are in favor of different behaviors for the same observation. If the input observation is extended with a condition\footnote{Conditions are not necessarily to be fed into the neural network at input layer with original observations, one can also concatenate them with intermediate features as needed, which is especially the case when image-like observations are used.} that exclusively embodies the relevant reward parameters, gradients will consist of two components during backpropagation. One drives the neural network to extract useful features from the original observation; the other drives the network to interpret different input conditions and combine them with the extracted features properly to generate specifically desirable behaviors as well as value estimations. With this intuition above, we expect such a conditional deep neural policy can learn to adapt its characteristics as the input condition changes.

We denoted a conditional policy parameterized by $\theta$ as $\pi_{\theta}\left(a|s,c\right)$ or $\pi_\theta^{c}$ for simplicity, where $c$ represents the input condition. Assume $q_c\left(s_{0}\right)$ is the initial state distribution under $c$, then the cDRL optimization target is formulated as:

\begin{equation}
    \mathcal{L}\left(\theta\right)=-\mathbb{E}_{s,a\sim q_c,\pi_{\theta}^c;c\sim \mathcal{C}^e}\left[\sum_{t=0}^{\infty}\gamma^{t}r_c\left(s_t,a_t,s_{t+1}\right)\right]
\end{equation}

Apparently, the desired training data is nested, both the condition subspace $\mathcal{C}^e$ and the consequent conditional example spaces should be sufficiently explored. This may seem extremely inefficient in the first place, but keep in mind that the feature extraction task is shared across all reward parameters while various reward parameters will induce observation distributions with better diversity which in turn is beneficial for this task. We assume that the main learning burden for deep RL algorithms lies in extracting efficient high-level features strongly correlated with decision making from raw input information. Then it's possible for cDRL to learn without requiring more examples or notably prolonged training compared to standard deep RL as long as its extra task of preference adjustment is relatively simple. Actually, we evaluate both our approach and baselines with equal amount of training in section 4. The experimental results support our assumption well.

On the other hand, to enhance exploration diversity in $\mathcal{R}_-^e$, we adopt the asynchronous methods as in [16] by running a batch of environments in parallel with individually sampled reward parameters which are updated periodically. Other measures of standard deep RL algorithms to improve data efficiency and learning stabilization are kept unchanged. We next describe the conditional versions of A3C [16], DDPG [17], and Deep Q-learning [2] in detail. As a general case, the full cDRL algorithm is outlined in Algorithm 1. For the chosen deep RL frameworks, any neural network approximator that takes observation as input will become conditional in the sense that original observation has been concatenated with an extra condition. We discuss these cDRL frameworks in detail as below.

\begin{itemize}
\item \textbf{Conditional A3C:} the algorithm maintains a conditional policy $\pi_{\theta}\left(a|s,c\right)$ and a conditional value estimation function $V_{\psi}^c\left(s\right)$. The policy is optimized according to the advantage-based policy gradient $A_c\left(s,a\right)\nabla_{\theta}\log\pi_{\theta}\left(a|s,c\right)$, where $A_c\left(s,a\right)$ is the conditional advantage of action $a$ for state $s$ under condition $c$. For moment $t$ and a following episode of length  $H$, $A_c\left(s_t,a_t\right)$ is estimated by $\sum_{k=0}^{H-1}\gamma^{k}r_{t+k}^c+\gamma^{H}V_{\psi}^c\left(s_{t+H}\right)-V_{\psi}^c\left(s_t\right)$. We adopt PPO [18] to stabilize learning. As an online RL algorithm, a big batch size of sampling environments is important for the success of conditional training.

\begin{algorithm}
    \caption{Conditional Deep Reinforcement Learning}
    \begin{algorithmic}[1]
        \REQUIRE $\mathcal{R}_{-}^e$: near-optimal reward parameter subspace
        \REQUIRE $\mathcal{M}$: affine transformation from $\mathcal{R}_-^e$ to $\mathcal{C}^e$
        \REQUIRE $\alpha$: step size hyperparameter
        \REQUIRE $\tau$: refresh period of reward parameters
        \STATE randomly initialize $\theta$, $step\leftarrow 1$
        \WHILE {not done}
        \STATE sample a batch of reward parameters from $\mathcal{R}_-^e$, apply them to environments separately, and project to $\mathcal{C}^e$ with $\mathcal{M}$ for output
        \WHILE {$step\%\tau\neq 0$}
        \STATE Evaluate $\nabla_{\theta}\mathcal{L}\left(\theta\right)$ with respect to a batch of mixed examples
        \STATE Update $\theta\leftarrow\theta-\alpha\nabla_{\theta}\mathcal{L}\left(\theta\right)$
        \STATE $step\leftarrow step+1$
        \ENDWHILE
        \ENDWHILE
    \end{algorithmic}
\end{algorithm}

\item \textbf{Conditional DDPG:} the algorithm learns a conditional Q-function $Q_{\psi}^c\left(s,a\right)$ with parameter $\psi$ by fitting the target value $r_c\left(s,a,s'\right)+\gamma Q_{\psi^{-}}^c\left(s',\pi_{\theta^{-}}^c\left(s'\right)\right)$ based on the bootstrapping property of Bellman Equation. $\pi_{\theta}^c$ is the conditional deterministic policy parameterized by $\theta$, which is optimized through the gradient directly stemming from $Q_{\psi}^c$, given by $\nabla_a Q_{\psi}^c\left(s,a\right)|_{a=\pi_{\theta}^c\left(s\right)}\nabla_{\theta}\pi_{\theta}^c\left(s\right)$. Independent target networks $Q_{\psi^{-}}^c$ and $\pi_{\theta^{-}}^c$ are used and softly updated with a temperature parameter to stabilize learning. Conditional transitions [$s\oplus c$, $a$, $s'\oplus c$, $r_c\left(s,a,s'\right)$] are pushed into a replay buffer and resampled in batches for training, where $\oplus$ represents concatenating operation.
\item \textbf{Conditional DQN:} similar to conditional DDPG, this algorithm also learns a conditional $Q_{\theta}^c\left(s,a\right)$ parameterized by $\theta$ from conditional transitions but with a different target value given by $r_c\left(s,a,s'\right)+\gamma\max_{a'\in\mathcal{A}}Q_{\theta^{-}}^c\left(s',a'\right)$, where $\mathcal{A}$ is a discrete collection of all available actions for state $s'$. A lagged target network $Q_{\theta^{-}}^c$ is used to stabilize learning, which is updated with a lower frequency than $Q_{\theta}^c$. For a certain state $s$ under condition $c$, the desired action is selected according to $Q_{\theta}^c\left(s,a\right)$ with $\varepsilon$-greedy strategy during training or greedy strategy at test time. No explicit conditional policy is learned in this algorithm.
\end{itemize}

\section{RELATED WORK}

As proof-of-concept researches, [7][8][9] used exhaustive search to examine the nature of reward functions and verify the benefits of well-designed rewards. [10] made one step towards pragmatic applications and proposed a lightweight approach which utilizes policy gradient to optimize reward parameters online. This approach requires an explicit model of the Markov Decision Process which is impractical for complex or continuous tasks. Another research direction adopts nested optimizations which apply a high-level reinforcer or genetic programmer to optimize reward parameters while optimizing RL policies [11][12][13][14]. These approaches are strongly bounded on task complexities and available computation resources. [15] presented a Bayes approach for reward design by estimating a posteriori over optimal rewards with parameter samples and their performance. They used alternative planning methods instead of RL to circumvent the intractability of the original idea which also confines their approach to relatively simple tasks.

Meta reinforcement learning (meta-RL) is a sort of RL algorithms designed for fast adaption to new tasks via learning internal representations broadly suitable to a certain task distribution. Theoretically, meta-RL could be trained to adapt to different reward parameters and perform similar hindsight policy characteristic tweaking as in cDRL. However, it relies on either special network structures [21][22][23] or a special loss computed by two consecutively sampled batches [24], which significantly increases learning difficulty and inevitably demands for longer training periods. Besides, sufficient meta-adaptions are needed before performance evaluations on reward functions can be executed. In contrast, as a highly specialized approach for hindsight reward tweaking, cDRL is efficient both in training and evaluation.

Analogies has been made between deep reinforcement learning with Actor-Critic structures and generative adversarial nets (GANs) [25][26]. Similarly, cDRL also corresponds to conditional GANs (cGANs) [27] for applying the same methodology: featured conditions are added to the input and trained to be sensitive for corresponding data distributions which in cGANs are manipulated via data feeding while in cDRL are determined by reward parameters in a relatively unstraightforward way. cGANs have been widely reported to sharpen the predictive distributions for both the discriminator and generator, thus significantly improve the visual quality of generated images [27][28]. They are also easier to train than vanilla GANs [29]. These facts endorse the effectiveness and feasibility of our approach in a way.

\section{CASE STUDY: HINDSIGHT POLICY BOOSTING}

\subsection{Method Formulation}

After a cDRL policy is trained, all learnable parameters are held constant as $\theta^*$. Given a certain fitness function, the input condition $c$ becomes the only control interface for further optimization. Then, search for the optimal policy $\pi_{\theta^*}^{c^*}$ reduces to search for the optimal input condition $c^*$, which is given by:

\begin{equation}
    c^*=\argmax_{c\in \mathcal{C}^s}\mathbb{E}\left[\mathcal{F}\left(\pi_{\theta^*}^c\right)\right]
\end{equation}

\noindent where $\mathcal{F}\left(\pi_{\theta^*}^c\right)$ represents any evaluation process on the conditional policy $\pi_{\theta^*}^c$ which returns scalar fitness scores. $\mathcal{C}^s\supseteq\mathcal{C}^e$ stands for the searching space for $c^*$ at test time. With sufficient training, a cDRL policy can generalize to boarder space within $\mathcal{C}$ such that the potential optimal condition may locate outside $\mathcal{C}^e$ within which the policy is trained. Since we have little prior knowledge about $\mathcal{C}^s$ with respect to agent performance, a natural choice for optimization method is genetic programming [20], which places no assumption on problem domains and ensures global optimum in searching space.

Compared to separate 'trial-train-test' circles, cDRL combines the first two phases into a single one-time training process, which makes the hindsight reward tweaking and evaluation much more flexible. This is the key advantage of cDRL over former ORP solutions on large-scale complex tasks which could take tremendous amount time of training [19].

\begin{figure*}[t]
\centering
\includegraphics[width=1.0\textwidth]{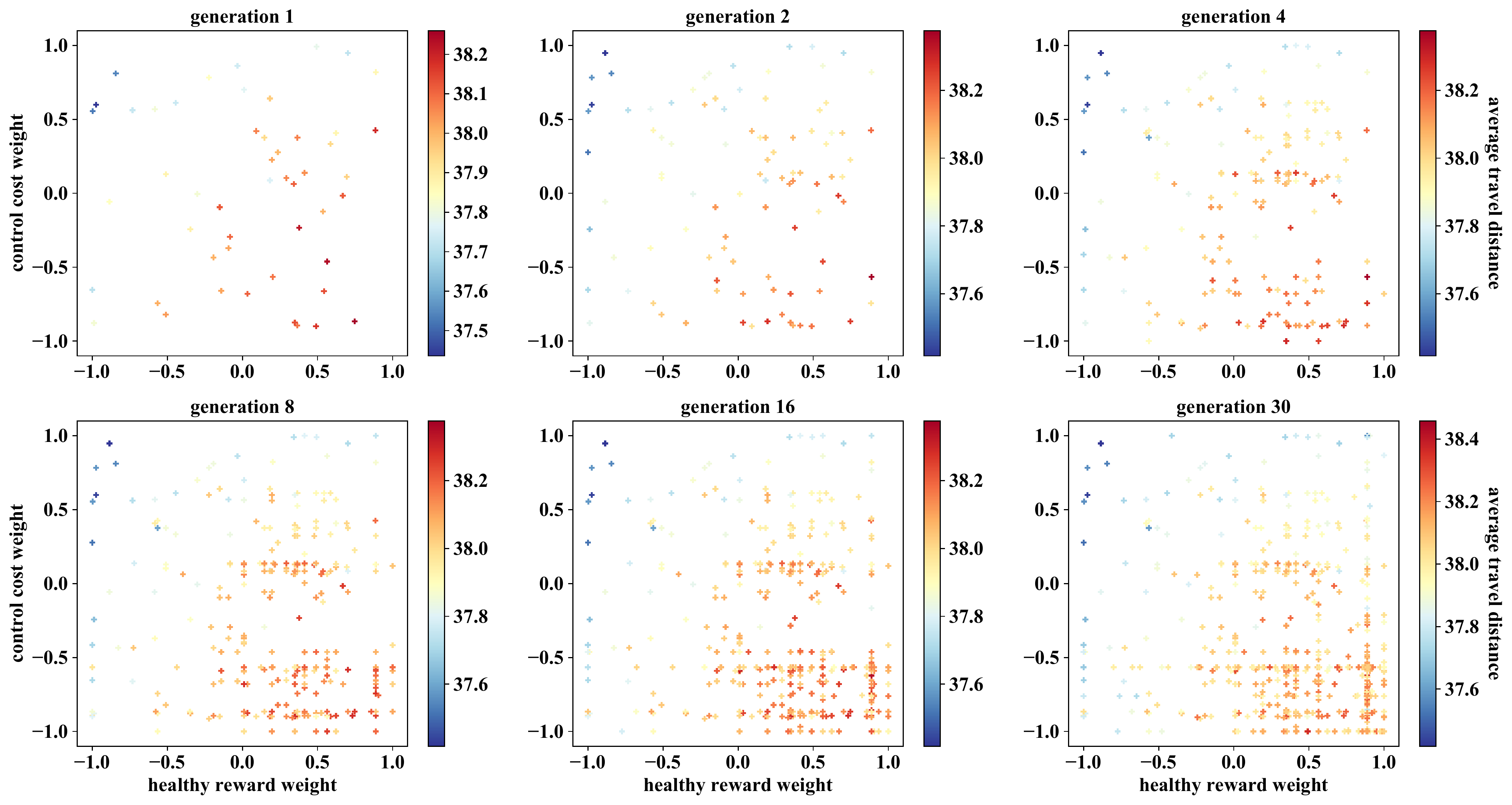} 
\caption{Evolution heatmaps of conditional A3C on Walker2d domain. Each cross represents an individual from some generation of size 50. Accumulated populations at generation 1, 2, 4, 8, 16, and 30 are illustrated. Displaying range of each dimension are limited within [-1,1] for better detail visibility. As expected, crosses gradually aggregate in regions corresponding to long average travel distances which indicate desirable interaction effects among different reward parameters. Note that there are multiple high-density regions which could likely be interpreted as local optima.}
\label{fig3}
\end{figure*}

\subsection{Experimental Configurations}

We choose MuJoCo [30] integrated in OpenAI Gym [31] to test the proposed cDRL approach. MuJoCo provides excellent physics simulations and is widely adopted for benchmarking high-dimensional continuous control tasks. Our target is to verify if a trained cDRL policy can be tweaked by the input condition and if a hindsight boosted cDRL policy outperforms policies trained by standard deep RL algorithms with default reward parameters given equal amount of training. In specific, we choose three locomotion tasks: HalfCheetah, Walker2d, and Ant, all of which aim to maximize the agents' forward velocity without falling to the ground (if applicable). We apply the conditional versions of both A3C and DDPG to these domains and use deep polices trained by corresponding standard frameworks with default reward parameters in Gym environment settings, which are already well optimized, as our baselines for benchmarking.

We follow the original indicator features defined in Gym such as forward reward, healthy reward, control cost and contact cost, etc. and then a vector of ones acts as baseline reward parameters. When applying cDRL algorithms, we choose forward reward as anchor with constant weight 1.0 and the other reward parameters varied within [$1-\epsilon$,$1+\epsilon$] where $\epsilon$ is a small positive value. We set $\epsilon$ = 0.2 for HalfCheetah and Walker2d while $\epsilon$ = 0.05 for Ant. Note that this is not necessarily the case for tasks of which indicator features are not well scaled and substantially heterogeneous ranges might need to be specified for these tasks. During training, baseline and cDRL models use almost identical configurations and hyperparameters except for three main differences: a) cDRL uses a slightly modified network architecture to admit the extra input condition; b) cDRL independently samples reward parameters from the predefined ranges and refresh them periodically for every environment while baseline uses default reward parameters for all environments; c) baseline models are trained 3$\times$10\textsuperscript{4} more agent steps than cDRL models to compensate their extra exposure to environments during hindsight optimization.

\textbf{A3C:} The Actor and Critic have identical and separated fully-connected network structures with 3 hidden layers of 256 units and tanh nonlinearity. We use PPO loss [18] to compute stabilized policy gradient with a clip range of 0.2 and Adam [32] to update network parameters with a learning rate of 3$\times$10\textsuperscript{-4}. 50 environments are run in parallel with an episode length of 2048, a full batch from all 50 environments are divided into 4 minibatches and utilized for 16 epochs per update. For each experiment, a total number of 810 updates are performed with samples of about 8$\times$10\textsuperscript{7} agent steps. We use 0.99 and 0.95 for discounting factor and truncation factor of generalized advantage estimation (GAE) [33] respectively. The entropy coefficient is set to 0.0 as default for MuJoCo tasks in Gym. For the conditional version of A3C, we resample the reward parameters for all environments every 10 updates.

\textbf{DDPG:} The Actor and Critic have identical and separated fully-connected network structures with 2 hidden layers of 64 units and ReLU nonlinearity after which layer normalization [34] is applied. Adaptive parameter noise is used for HalfCheetah and Walker2d for exploration while Ornstein-Uhlenbeck noise [17] is used for Ant. 50 environments are run in parallel to sample transitions which are pushed into a replay buffer of size 1$\times$10\textsuperscript{6}. A batch of 128 transitions are resampled from the buffer for gradient calculation per update. Adam is used to update network parameters with learning rates 1$\times$10\textsuperscript{-4} and 1$\times$10\textsuperscript{-3} for Actor and Critic respectively. For each experiment, 50 updates are performed after every 100 agent steps which is repeated for 2$\times$10\textsuperscript{4} times. We use a discounting factor of 0.99 and a soft update coefficient of 0.001 for target networks. For the conditional version of DDPG, we resample the reward parameters every 2048 agent steps.

For hindsight optimization, we choose the average travel distance of 1000 continuous steps along forward direction for 50 random seeds as the fitness function. Outliers caused by occasional fallings are excluded for stabilization while runs with over 10 fallings return 0. Genetic programming with real-valued encoding was applied to a predefined condition subspace $\mathcal{C}^s$ of which each dimension varies within [-2,2] for conditional A3C policies while [-10,10] for conditional DDPG polices. A population size of 50 is used for every generation. We use tournament selection strategy with elite preservation and single point crossover with a probability of 0.8 for recombination after which a mutation could happen with probability 0.1. All optimizations evolve for 30 generations and the extra environment exposures during this process are compensated for baseline trainings as mentioned above.

\begin{figure*}[t]
\centering
\includegraphics[width=1.0\textwidth]{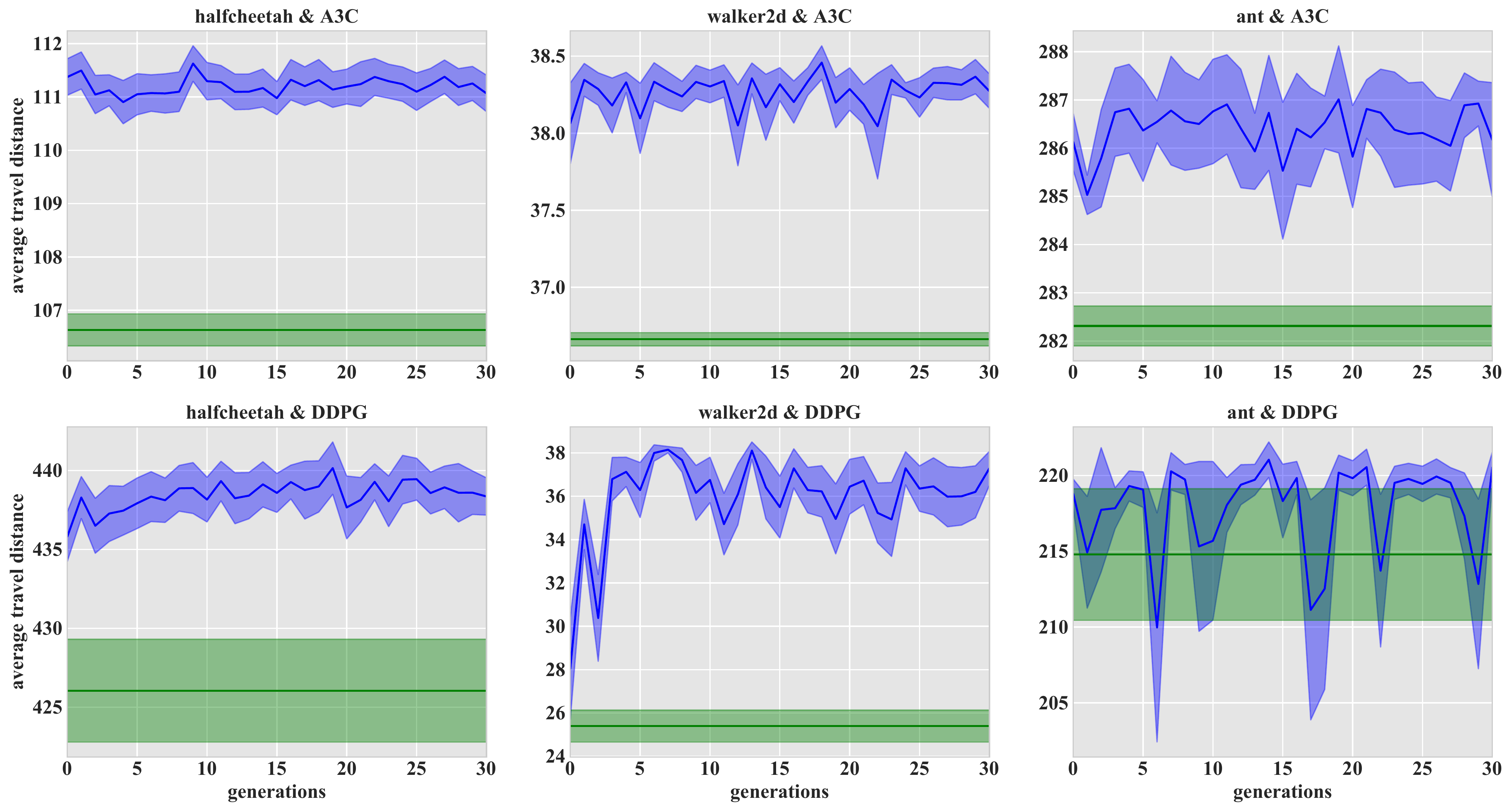} 
\caption{Performance comparisons between cDRL (blue) and baseline (green) policies. Average forward distance of 1000 steps out of 50 trials are evaluated as reference and used as fitness when applying genetic programming on condition spaces of cDRL policies. As illustrated, the top elite cDRL policies during evolution exhibit considerably superior performance to baseline policies.}
\label{fig4}
\end{figure*}

\subsection{Results}

The evolution processes are visualized in Fig. 3. We stored the populations of all generations during hindsight optimization of the conditional A3C policy trained on Walker2d domain which has 2 non-anchor reward parameters. These 2D parameter coordinates are illustrated with scatter plotting of which colors indicate measured performance and densities reveal evolution trends. Obviously, a cDRL policy exhibits distinct characteristics, and thus different performance, while the input condition is varying in the condition space, which validates the feasibility of cDRL as we expected. From the evolution heatmaps in Fig. 3, we can also learn some good intuition about the interaction mechanisms of reward parameters for Walker2d domain: bigger healthy reward plus smaller control cost tend to result in better ability of running forward.

To verify the ability of cDRL in boosting policy performance, we compared the best individuals of generations during genetic evolution with baseline models. As shown in Fig. 4, hindsight optimized cDRL policies consistently yields longer travel distance than baseline policies in all the three domains, which proves that the long-period influences of reward parameters can not only be modeled by cDRL but also further utilized to search for better polices. Apparently, one can also make use of cDRL approach to achieve better performance on real-world RL tasks where hand-designed raw reward parameters works but sophisticated interaction mechanisms exists among them.

For better understanding of cDRL, we performed extra qualitative experiments on its unique hyperparameters, i.e. refresh period $\tau$ and exploration range $\epsilon$ of reward parameters. We found that a too small $\tau$ or too big $\epsilon$ would lead to significant performance decrease. The former is caused by unbalanced sample structure which overemphasizes horizontal diversity in reward parameter space at the expense of vertical data sufficiency in conditional example spaces; the latter results from the loss of focus on core near-optimal region. In practice, the selection of these two hyperparameters, especially for $\epsilon$, is dependent on specific task properties. As a rule of thumb, one should use as big a batch of sampling environments as possible and avoid very small $\tau$s. For unfamiliar task domains, one should start with small exploration ranges for each reward parameter given a group of rewards that already works.

\section{CONCLUSION AND FUTURE WORK}

We observe the fundamental role of reward design in RL, refer to the wisdom of 'conditional deep learning', and propose a new paradigm for deep RL called cDRL which models the influences of reward functions while doing its original job. This approach is scalable for modern complex RL tasks. We successfully verify the feasibility of cDRL with several experiments on MuJoCo tasks and demonstrate one potential application in hindsight performance boosting of trained policies. Our approach tries to bridge the gap between reward changes and their actual effects by exempting routine trainings and enabling hindsight reward tweaking with more handy feedbacks. Importantly, cDRL doesn't require substantial modifications on learning processes of standard deep RL frameworks.

Essentially, our trained conditional polices provides a manipulation interface stemming from the learned long-period functioning mechanism of several key factors (reward parameters in this paper) in the form of internal neural weights. Sensitivity takes no less credit than modeling accuracy does for the effectiveness of cDRL. Given the success of our approach on reward parameters, it's promising to extend this paradigm to other hyperparameters with fundamental but delayed influences, such as the discounting factor, truncation factor of GAE, etc. as long as they could be individually configured in separate sampling pipelines. We'll leave this research for future work.

\addtolength{\textheight}{-6.85cm}   





\end{document}